\documentclass[letterpaper, 10 pt, conference]{ieeeconf}

\IEEEoverridecommandlockouts
\usepackage{afterpage}
\usepackage{xspace}
\usepackage{amssymb,paralist,epsfig,standalone,bm,placeins}  % assumes amsmath package installed
\usepackage{graphicx} % for pdf, bitmapped graphics files
\usepackage{xcolor}
\usepackage{multirow}
\usepackage{makecell}
\usepackage{multicol}
\usepackage[utf8]{inputenc}
\usepackage{amsmath,amsfonts,booktabs,cite} % general useful stuff
\usepackage{siunitx,textcomp} % for the degree symbol
\usepackage[hidelinks]{hyperref} % to have hyperlinks and for the \ref macros
\let\labelindent\relax
\usepackage[inline]{enumitem} % inline enumerate
\usepackage[nolist]{acronym} % acronyms by the \ac{label} command
\usepackage{caption}
\usepackage{tikz,standalone,pgfplots}
\usetikzlibrary{patterns}
\usepackage{pgfplots, pgfplotstable}
\graphicspath{{figures_tex/}}

\usepackage{changes}
\usepackage[ruled,vlined]{algorithm2e}

\usepackage{subcaption}
\usepackage{fp}
\usepackage{forloop}
\usepackage{siunitx} %for printing
\usepackage{contour}
\usepackage{color}
\usepackage{booktabs} % for professional tables
\usepackage{multirow} % For merging multiple rows
\usepackage{mathtools} %for multlined

\makeatletter
\newcommand{\removelatexerror}{\let\@latex@error\@gobble}
\makeatother
\pgfplotsset{compat=1.12}
\newacro{rl}[RL]{Reinforcement Learning}
\newacro{il}[IL]{Imitation Learning}
\newacro{dmp}[DMP]{Dynamic Movement Primitive}
\newacro{dof}[DoF]{Degree of Freedom}
\acrodefplural{dof}[DoFs]{Degrees of Freedom}
\newacro{pca}[PCA]{Principal Component Analysis}
\newacro{ik}[IK]{Inverse Kinematics}
\newacro{bilbo}[BILBO]{Bimanual dynamic manipulation using Imitation Learning for Bag Opening}

\newcommand{\figref}[1]{\hyperref[#1]{Fig.~\ref*{#1}}}
\newcommand{\tabref}[1]{\hyperref[#1]{Table~\ref*{#1}}}
\newcommand{\secref}[1]{\hyperref[#1]{Section~\ref*{#1}}}
\newcommand{\algoref}[1]{\hyperref[#1]{Algorithm~\ref*{#1}}}

\def\P{\mathbf{P}}

\def\p{\mathbf{p}}

\def\s{\mathbf{s}}

\definecolor{findOptimalPartition}{HTML}{D7191C}
\definecolor{storeClusterComponent}{HTML}{FDAE61}
\definecolor{dbscan}{HTML}{ABDDA4}
\definecolor{constructCluster}{HTML}{2B83BA}

\title{\LARGE \bf
Dynamic Manipulation of Deformable Objects using \\ Imitation Learning with Adaptation to Hardware Constraints
}

\author{Eric~Hannus$^{1}$, Tran~Nguyen~Le$^{1}$, David Blanco-Mulero$^{1,2}$, Ville~Kyrki$^{1}$%
\thanks{This work was financially supported by Business Finland (decision 9249/31/2021) and European Union's Horizon Europe programme project SoftEnable (grant agreement No. 101070600).} \thanks{$^{1}$ Intelligent Robotics Group at the Department of Electrical Engineering and
Automation, School of Electrical Engineering, Aalto University, Espoo, Finland.
\texttt{\{firstname.lastname\}{@}aalto.fi}}
\thanks{$^{2}$ Institut de Robòtica i Informàtica Industrial, CSIC-UPC, Barcelona, Spain. \texttt{david.blanco.mulero@upc.edu}}
}

\makeatletter
\let\@oldmaketitle\@maketitle% Store \@maketitle
\renewcommand{\@maketitle}{\@oldmaketitle% Update \@maketitle to insert...
  \setcounter{figure}{0}
    \vspace{1.5cm}
    \centering
    \def\svgwidth{\linewidth}
    \fontsize{6}{6}%\selectfont\sf
\begingroup%
  \makeatletter%
  \providecommand\color[2][]{%
    \errmessage{(Inkscape) Color is used for the text in Inkscape, but the package 'color.sty' is not loaded}%
    \renewcommand\color[2][]{}%
  }%
  \providecommand\transparent[1]{%
    \errmessage{(Inkscape) Transparency is used (non-zero) for the text in Inkscape, but the package 'transparent.sty' is not loaded}%
    \renewcommand\transparent[1]{}%
  }%
  \providecommand\rotatebox[2]{#2}%
  \newcommand*\fsize{\dimexpr\f@size pt\relax}%
  \newcommand*\lineheight[1]{\fontsize{\fsize}{#1\fsize}\selectfont}%
  \ifx\svgwidth\undefined%
    \setlength{\unitlength}{1152bp}%
    \ifx\svgscale\undefined%
      \relax%
    \else%
      \setlength{\unitlength}{\unitlength * \real{\svgscale}}%
    \fi%
  \else%
    \setlength{\unitlength}{\svgwidth}%
  \fi%
  \global\let\svgwidth\undefined%
  \global\let\svgscale\undefined%
  \makeatother%
  \begin{picture}(1,0.2)%
    \lineheight{1}%
    \put(0,0){\includegraphics[width=\linewidth,page=1]{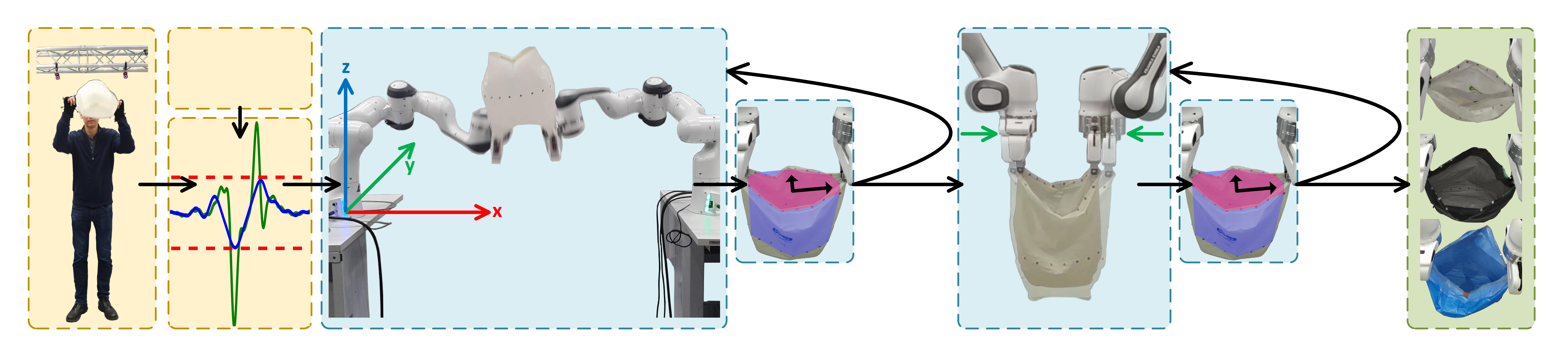}}%

    \fontsize{7.5pt}{7.5pt*1.2}

    \put(0.005,0.24){\color[rgb]{0,0,0}\makebox(0,0)[lt]{\lineheight{1.25}\smash{\begin{tabular}[t]{c}  Single \\ demonstration \end{tabular}}}}%

    \put(0.155,0.24){\color[rgb]{0,0,0}\makebox(0,0)[ct]{\lineheight{1.25}\smash{\begin{tabular}[t]{c} Robot kinematic \\ constraints \end{tabular}}}}%

    \put(0.154,0.00){\color[rgb]{0,0,0}\makebox(0,0)[ct]{\lineheight{1.25}\smash{\begin{tabular}[t]{c} Constrained DMP \end{tabular}}}}%

    \put(0.200,0.22){\color[rgb]{0,0,0}\makebox(0,0)[lt]{\lineheight{1.25}\smash{\begin{tabular}[t]{c} Learned dynamic manipulation primitive \end{tabular}}}}%

    \put(0.508,0.045){\color[rgb]{0,0,0}\makebox(0,0)[ct]{\lineheight{1.25}\smash{\begin{tabular}[t]{c} Compute task \\ performance \end{tabular}}}}%
    
    \put(0.576,0.095){\color[rgb]{0,0,0}\makebox(0,0)[ct]{\lineheight{1.25}\smash{\begin{tabular}[t]{c} Sufficient \end{tabular}}}}%

    \put(0.570,0.18){\color[rgb]{0,0,0}\makebox(0,0)[ct]{\lineheight{1.25}\smash{\begin{tabular}[t]{c} Insufficient \end{tabular}}}}%

    \put(0.573,0.22){\color[rgb]{0,0,0}\makebox(0,0)[lt]{\lineheight{1.25}\smash{\begin{tabular}[t]{c} Quasi-static refinement motions \end{tabular}}}}%

    \put(0.790,0.045){\color[rgb]{0,0,0}\makebox(0,0)[ct]{\lineheight{1.25}\smash{\begin{tabular}[t]{c} Compute \\updated  task \\ performance \end{tabular}}}}%

    \put(0.860,0.095){\color[rgb]{0,0,0}\makebox(0,0)[ct]{\lineheight{1.25}\smash{\begin{tabular}[t]{c} Sufficient \end{tabular}}}}%

    \put(0.860,0.18){\color[rgb]{0,0,0}\makebox(0,0)[ct]{\lineheight{1.25}\smash{\begin{tabular}[t]{c} Insufficient \end{tabular}}}}%

    \put(0.894,0.00){\color[rgb]{0,0,0}\makebox(0,0)[lt]{\lineheight{1.25}\smash{\begin{tabular}[t]{c} Final states \end{tabular}}}}%
    \fontsize{7.0pt}{7.0pt*1.2}
    \put(0.152,0.184){\color[rgb]{0,0,0}\makebox(0,0)[ct]{\lineheight{1.25}\smash{\begin{tabular}[t]{c}
    $
    \begin{aligned}
    &\underline{q} \leq  q \leq \overline{q} \\[-4pt]
    &\underline{\dot{q}} \leq  \dot{q} \leq \overline{\dot{q}} \\[-4pt]
    &\underline{\ddot{q}} \leq  \ddot{q} \leq \overline{\ddot{q}}
    \end{aligned}
    $
    \end{tabular}}}}%

  \end{picture}%
\endgroup%

  \captionof{figure}{Our framework combines a dynamic motion primitive that adheres to the robots' constraints, for quick task progression, with quasi-static motions for final adjustments. 
  We evaluate the framework by implementing \textit{BILBO: Bimanual dynamic manipulation using Imitation Learning for Bag Opening}, a system for a practical bag-opening task. 
 \label{fig:full_pipeline} 
}}
\makeatother

\begin{document}
\maketitle
\thispagestyle{empty}
\pagestyle{empty}

%%%%%%%%%%%%%%%%%%%%%%%%%%%%%%%%%%%%%%%%%%%%%%%%%%%%%%%%%%%%%%%%%%%%%%%%%%%%%%%%

\begin{abstract}

Imitation Learning (IL) is a promising paradigm for learning dynamic manipulation of deformable objects since it does not depend on difficult-to-create accurate simulations of such objects. 
However, the translation of motions demonstrated by a human to a robot is a challenge for IL, due to differences in the embodiments and the robot's physical limits. These limits are especially relevant in dynamic manipulation where high velocities and accelerations are typical.
To address this problem, we propose a framework that first maps a dynamic demonstration into a motion that respects the robot's constraints using a  constrained Dynamic Movement Primitive. Second, the resulting object state is further optimized by quasi-static refinement motions to optimize task performance metrics. This allows both efficiently altering the object state by dynamic motions and stable small-scale refinements. 
We evaluate the framework in the challenging task of bag opening, designing the system \textit{BILBO: Bimanual dynamic manipulation using Imitation Learning for Bag Opening}.
Our results show that BILBO can successfully open a wide range of crumpled bags, using a demonstration with a single bag.
See supplementary material at \url{https://sites.google.com/view/bilbo-bag}.

\end{abstract}

%%%%%%%% Start of document

\section{Introduction}
\label{sec:introduction}
Dynamic manipulation has shown great potential, particularly in the manipulation of deformable objects~\cite{hietala2022learning, xu2022dextairity, gu2023shakingbot}. Unlike quasi-static manipulation like pick-and-place, dynamic manipulation utilizes forces of acceleration for the success of the task~\cite{DynamicMason93}. Therefore dynamic actions can achieve object configurations that are not reachable through quasi-static manipulation \cite{ha2022flingbot, xu2022dextairity}.
For instance, to dynamically manipulate objects such as cloths, a typical approach is to learn the skill in simulation and then transfer it to the real world~\cite{hietala2022learning, ha2022flingbot}.
However, simulation approaches can be unsuitable for more complex deformable objects such as bags, due to their 3D structure and the complex forces, for example, those related to aerodynamics \cite{xu2022dextairity}.
One solution to bypass this challenge is to learn the manipulation skills directly from human demonstrations using \ac{il}.

\ac{il} is a learning paradigm for teaching skills to robots by providing demonstrations \cite{argall2009survey}.
The demanding dynamics of dynamic manipulation including high accelerations and velocities highlight the \ac{il} challenge known as the correspondence problem~\cite{NehanivCorrespondenceProblem2001}.
Specifically, intuitive dynamic motions exhibited by human demonstrators may be impossible to transfer directly to robots, due to  differences between the bodies and constraints of the robot actuation.

To overcome this problem, we propose a framework, shown in Fig. \ref{fig:full_pipeline}, that first transforms a dynamic human demonstration into a motion that adheres to the robot constraints by utilizing constrained \acp{dmp}~\cite{dahlin2021temporal, sidiropoulos2023novel}.
In the second stage of the framework, quasi-static motions are used for stable refinement of the deformable object state.
Our framework decides whether to repeat an action or proceed to the next stage using the task performance metrics.
These task specific metrics can be computed utilizing the same system used for recording the demonstration, e.g. a motion capture system.

We demonstrate the benefits of the proposed framework in the task of bimanual bag opening with the novel system \textit{BILBO: Bimanual dynamic manipulation using Imitation
Learning for Bag Opening}. \acused{bilbo}\ac{bilbo} employs a dynamic motion to optimize the volume and area of the bags, followed by a linear refinement motion to enhance the opening roundness~(see Fig.~\ref{fig:full_pipeline}). 
Our experiments show that a single human demonstration is sufficient for BILBO to successfully open bags of a wide range of sizes and material properties, even when they are initialized in a highly crumpled state. 
The experimental results additionally emphasize 
the importance of prioritizing high velocities and accelerations in dynamic manipulation over strictly following the demonstrated path.

In summary, our contributions include:
\begin{itemize}
    \item A novel \ac{il} framework for learning dynamic manipulation of deformable objects with adaptation to hardware constraints using constrained \acp{dmp} and refining the object state using quasi-static motions.
    \item An extensive evaluation of three constrained DMP candidates in a dynamic manipulation setting.
    \item A new definition of a bag volume metric, and a rim area metric based on $\alpha$-shapes, for the bag-opening task. 
    \item  A thorough empirical evaluation of BILBO's performance and generalization capabilities on a wide range of bags made of different materials, sizes, and shapes.

\end{itemize}

\section{Related work}
\label{sec:related_work}
\subsection{Learning Dynamic Manipulation of Deformable Objects}
\label{sec: LearnedDynamic}

Dynamic manipulation of deformable objects has been studied for 1D ropes \cite{chi2022iterative}, 2D cloths \cite{ha2022flingbot, QDP2023, chi2022iterative, hietala2022learning, balaguer2011combining} and 3D objects like pizza dough \cite{satici2016pizza}. These methods typically utilize simulations\cite{chi2022iterative, zhang2021lostarc, ha2022flingbot, QDP2023, hietala2022learning, satici2016pizza} thanks to the recent advancement in soft-body physics engines, making it feasible to gather large amounts of data.
While prior work has successfully transferred dynamic manipulation policies learned in simulation to the real-world~\cite{ha2022flingbot, hietala2022learning, QDP2023}, there is a noticeable reality gap in dynamic tasks~\cite{blancomulero_2024_benchmarking_cloth_manip}.
This gap is even more pronounced in bag manipulation, where no simulation engine is currently capable of accurately simulating plastic bags~\cite{gao2023iterative}.
To begin with, physic engines require modeling the complex aerodynamics effects that take place in the real world when dynamically manipulating bags~\cite{xu2022dextairity}.
Furthermore, the perception of plastic bags is challenging due to their translucent and reflective material \cite{chen2023autobag, gao2023iterative}, which are difficult to capture in simulation to ensure successful sim-to-real transfer. 
Although prior work has demonstrated manipulation of bags in simulation~\cite{seita2021learning, antonova2021DEDO},
they provide no guarantees of a successful sim-to-real transfer.

\subsection{Bag Manipulation}
\label{sec: BagManipulation}
The research on bag manipulation has mainly focused on two perspectives.
The first direction has focused on learning the dynamics of bags~\cite{weng2021graphbased, antonova2021sequential}, where the problem of learning to manipulate the bags is neglected. The second direction has used pre-defined primitives for manipulating the bag.
One option for achieving this is to leverage self-supervision methods, where the bags can be marked and segmented, to learn the perception of the bag state, which is used in a pre-defined reasoning scheme to apply pre-defined manipulation primitives~\cite{chen2023autobag, chen2023SLIPbagging, gu2023shakingbot}.
These works dealt with the bagging problem, that is, opening a bag, inserting items, and lifting it \cite{chen2023autobag}. Other works have studied learning where to direct pick-and-place actions ~\cite{gao2023iterative, bahety2022bag, seita2021learning}.
While many works rely on quasi-static actions~\cite{gao2023iterative, bahety2022bag, seita2021learning}, some recent works have demonstrated the effectiveness of dynamic manipulation \cite{xu2022dextairity, gu2023shakingbot}.
In~\cite{xu2022dextairity}, the authors proposed a method to open bags by applying an air stream using a blower.
Rather than executing pre-defined motions, or utilizing unconventional actuators, in this work we learn a dynamic motion from a single human demonstration, which can effectively open bags using standard grippers.

Another alternative to learn motions that are not pre-defined is using \ac{rl} to train a policy in a simulated environment~\cite{antonova2021DEDO}. However, standard model-free algorithms are unable to succeed in bag manipulation tasks~\cite{antonova2021DEDO}. Due to the aforementioned limitations of bag simulation, and the poor performance of standard \ac{rl} algorithms for bag manipulation tasks, BILBO uses \ac{il} to circumvent these limitations and apply dynamic manipulation to the problem of bag opening. Specifically, we utilize the \ac{dmp} framework \cite{ijspeert2013dynamical} to learn directly from a single human demonstration in the real world, and bypass the perception issues discussed, by tracking robust hand poses rather than markers attached to the bag.

\section{Background}
\label{sec:background}
In this section, we introduce the formulation of \acp{dmp}~\cite{ijspeert2013dynamical}.
\acp{dmp} is a framework for encoding motions in a non-linear dynamic system. For point-to-point trajectories, \acp{dmp} represent the trajectories for a single \ac{dof} by the system of differential equations:
\begin{align} 
\tau \dot{z} &= \alpha_{z}(\beta_{z}(g-y)-z) + f(x),  \label{eq:dDMP1} \\
\tau \dot{y} &= z,  \label{eq:dDMP2}  \\
\tau \dot{x} &= -\alpha_{x}x.  \label{eq:dDMP3} 
\end{align}
The transformation system, expressed by equations \eqref{eq:dDMP1} and \eqref{eq:dDMP2}, resembles a mass-spring-damper system augmented with a forcing term $f(x)$ and a time constant $\tau$ for scaling the duration of the motion.
The equilibrium for the position $y$ in the system is given by $g$, which thus encodes the goal of the motion.
The parameters $\alpha_{z}$ and $\beta_{z}$ are positive constants tuned to achieve critical damping in the system when the forcing term is excluded.

The equation~\eqref{eq:dDMP3}, called the canonical system, describes the decay of the phase $x$, which replaces the explicit time dependence via a positive constant $\alpha_x$. This implicit representation of time enables synchronization between multiple \acp{dof} or with external systems \cite{ijspeert2013dynamical}, and makes it possible to slow down or stop the motion in the presence of tracking errors \cite{ijspeert2002movement %,saveriano2021dynamic
}.
By augmenting the forcing term $f(x)$ complex trajectories can be encoded in the system.
The forcing term is calculated as a weighted sum of $H$ exponential kernels:
\begin{align} 
f(x) &= \frac{\sum_{i=1}^{H}\Psi_i(x)w_i}{\sum_{i=1}^{H}\Psi_i(x)}x(g-y_0),  \label{eq:dDMP4}  \\
\text{where } \Psi_i(x) &= \text{exp} \left(-\frac{1}{2\sigma_i^2}(x-c_i)^2\right),  \label{eq:dDMP5} 
\end{align}
where $\sigma_i$ and $c_i$ define the widths and centers of the kernels, and $w_i$ are the weights.
The kernels are spaced exponentially in phase and, therefore, evenly in time \cite{ijspeert2013dynamical}. Fitting the weights is a supervised learning problem that can be solved using any function approximator. For systems with multiple \acp{dof}, the \acp{dmp} of each \ac{dof} share one canonical system, and only the transformation systems and forcing terms are unique for each \ac{dmp} \cite{ijspeert2013dynamical}.

\section{IL Framework for Dynamic Manipulation}
\label{sec: Framework}
The proposed framework, shown in Fig. \ref{fig:full_pipeline}, consists of two main components: the dynamic primitive encoded with constrained \acp{dmp}, and the quasi-static refinement. 
The dynamic motions induce a drastic change in the state of a deformable object, which is useful for making fast progress in a task.
However, this feature of dynamic manipulation also makes it ill-suited for fine-tuning the object state. This motivates the second stage of our framework, where repeated small-scale changes from quasi-static motions are used to gently refine the state.
After each stage, the object state is estimated to compute task performance metrics that are used to decide wheter to repeat the stage or proceed to the subsequent step.
As the aim of the refinement stage is to improve the result of the dynamic stage, the performance metrics should have stricter requirements for deeming the state sufficient after refinement.

\subsection{Adaptation to Constraints with Constrained DMPs}
\label{sec: ConstrainedDMPs}

Given a set of $N$ Cartesian poses with quaternion orientation $\mathbf{P} \in \mathbb{R}^{7 \times N}$ provided as a path in the robot workspace, and robot position constraints, we rely on the assumption that an \ac{ik} solver can produce feasible joint-space targets $\mathbf{Q} \in \mathbb{R}^{D \times N}$, denoted as:
\begin{equation}
    \mathbf{Q} = \text{IK}(\mathbf{P}_, \underline{\mathbf{q}}, \overline{\mathbf{q}} ),
\end{equation}
where $\underline{\mathbf{q}} \in \mathbb{R}^D$ and $\overline{\mathbf{q}} \in \mathbb{R}^D$ are the lower and upper joint position limits, respectively, and $D$ is the number of \acp{dof} of the robot.
The corresponding velocities $\mathbf{\dot{Q}}$ and accelerations $\mathbf{\ddot{Q}}$ can be calculated using the timestamps of the provided path.
Then, given $\mathbf{Q}$, $\mathbf{\dot{Q}}$, $\mathbf{\ddot{Q}}$, and the kinematic constraints denoted with lower and upper bars, the constrained joint-space trajectory $\mathbf{Q}^*, \mathbf{\dot{Q}}^*, \mathbf{\ddot{Q}}^* \in \mathbb{R}^{D \times M}$ can be computed with a constrained \ac{dmp} (CDMP):
\begin{equation}
    \mathbf{Q}^*, \mathbf{\dot{Q}}^*, \mathbf{\ddot{Q}}^* = \text{CDMP}(\mathbf{Q}, \mathbf{\dot{Q}}, \mathbf{\ddot{Q}}, 
    \underline{\mathbf{q}}, \overline{\mathbf{q}}, 
    \underline{\mathbf{\dot{q}}}, \overline{\mathbf{\dot{q}}},
    \underline{\mathbf{\ddot{q}}}, \overline{\mathbf{\ddot{q}}}
    ).
\end{equation}
Note that the dimensions of the constrained trajectory matrices may differ from the unconstrained ones, due to augmented duration, velocity, and acceleration.

The original DMP formulation described in Section~\ref{sec:background} does not support the encoding of constraints.
Here, we present three alternatives that extend \acp{dmp} with this functionality and can be used for dynamic manipulation. To the best of our knowledge, our work is the first to use constrained DMPs for dynamic manipulation of deformable objects.

One straightforward solution is to adjust the time constant $\tau$ of the \ac{dmp}, which we refer to as tau-\ac{dmp}.
This solution retains the path shape that a non-constrained DMP would produce, and enforces velocity and acceleration constraints by gradually increasing the value of $\tau$ to uniformly slow down the motion until the constraints are satisfied. Consequently, the method can only be used in offline settings.

Recently, more advanced modifications have been presented that are capable of adapting the trajectory to the constraints online \cite{dahlin2021temporal, sidiropoulos2023novel} and encode additional constraints like via-points and obstacles \cite{sidiropoulos2023novel}. 
The constrained \ac{dmp} formulation in \cite{dahlin2021temporal} is provided using temporal coupling, that is, online scaling of $\tau$ to proactively scale it before the velocity and acceleration limits are exceeded. We refer to this method as TC-DMP.
One downside of this method is that it does not ensure that acceleration constraints are guaranteed in the online setting presented in~\cite{dahlin2021temporal}.
In offline settings, the acceleration constraints can be guaranteed by appropriate tuning of the model parameters via trial-and-error.

The third method we consider formulates the problem of fitting the weights of a \ac{dmp} with constraints as an optimization problem~\cite{sidiropoulos2023novel}, which we refer to as Opt-\ac{dmp}.
In order to express position, velocity, and acceleration at any time point as affine functions in the \ac{dmp} weights, Opt-\ac{dmp} follows the equations from~\cite{sidiropoulos2021reversible}, thus removing the need for explicit integration. 
This makes it possible to formulate Opt-\ac{dmp} as a solution to a Quadratic Program.
As the constraints are no longer enforced by scaling $\tau$ the generated path might differ from the demonstration. Therefore, position constraints are explicitly part of the optimization problem in addition to the velocity and acceleration constraints.

\subsection{Quasi-Static Refinement Motions}
\label{sec: FRAMEWORK Refinement Motion}

In contrast to dynamic motions which affect the entire state of an object, quasi-static motions can be designed to fine-tune a deformable object state via local, small, directed changes.
We consider quasi-static motions that are reversible,  while the effects of dynamic manipulation can typically not be undone. Formally, we assume that for every action $a$ there exists a reverse action $a^-$ such that the state dynamics $\s'=f(\s,a)$ follow $f(f(\s,a),a^-)=\s$. Furthermore, we assume the actions to cause only small changes in the object state $\|\s - f(\s,a)\|<\delta_1, \forall a$.
Thus, the reversible action space makes quasi-static motions suitable for stable refinement of the state.

To cast the refinement as an optimization problem, we assume there exists a continuous cost function $C(\s)$ that measures the quality of the state. 
Thus, we want to incrementally refine the state by applying actions such that the cost function is minimized:
\begin{equation}
    a = \arg\min_a C(\s').
\label{eq:argmin_cost}
\end{equation}
Assuming the cost function is convex within the region of attraction given by the initial state of the local refinement phase, continuing the local refinements allows converging to a minimum. This follows from the fact that any action will have a bounded effect on the cost $\|\s - f(\s,a)\|<\delta_1 \Rightarrow |C(\s)-C(\s')|<\delta_2$.

\section{
BILBO: Bimanual dynamic manipulation using Imitation Learning for Bag Opening}
\label{sec:method}
We present BILBO, a system that uses the proposed framework for
opening bags via dynamic manipulation using a bimanual system, as depicted in Fig.~\ref{fig:full_pipeline}. Here, we assume that non-dynamic steps such as grasping and item insertion can be solved by existing methods~\cite{chen2023autobag, gu2023shakingbot, chen2023SLIPbagging}.
First, we start by gathering a single human demonstration of a dynamic motion using a motion capture system.
Then, we utilize the aforementioned constrained \ac{dmp} approaches to augment the joint-space trajectory corresponding to the demonstration, so that it adheres to the robots' constraints.
The learned constrained motion is used to dynamically improve the volume and area of the bag.
As the ideal bag state should not only be expanded but also have a round opening, we design a quasi-static refinement motion for improving the rim elongation.
For evaluating the bag-opening performance and deciding whether the learned dynamic manipulation primitive or the linear refinement motion needs to be applied, we estimate the bag state using a motion capture system, and extract area, volume, and elongation metrics for the bag.

\subsection{Learning a Dynamic Motion from Human Demonstration}
\label{sec: Demo}
The human demonstration is captured using a motion capture system that tracks the hands of a human. The motion range of the demonstration should be within the robot's reach, or its amplitude must be scaled down.
At each time step of the demonstration, the distance between the hands is calculated, as a percentage of the maximum observed distance,  and used to define the desired distance between the robot grippers when replaying the motion for bags of different widths.
Any rotation that is not around the main rotation axis of the motion is filtered out. Then, the trajectory is smoothed to account for noise in the captured trajectory.
Finally, the trajectory is converted to joint space to apply the constrained \ac{dmp} approaches described in Section~\ref{sec: ConstrainedDMPs}.

% \subsection{Bag Metrics}
\subsection{Evaluating the Task Performance}
\label{sec: BagMetrics}

First, to define the performance metrics for the bag-opening task, we specify how to measure the state of the bag and its regions of interest.
To perceive the state, we employ a perception scheme based on reflective markers and a motion capture system.
Here, we capture the bag state after either the dynamic or refinement motion have finished,
which is more straightforward than tracking the bag state during its motion, especially if the motion is dynamic.

Each bag has two regions of reflective markers, one around the rim and another near the bottom. In addition, the inside of the rim has markers to facilitate detecting the rim even if it is folded. The marker points of the bag are defined as $\P =\{ \p_1, \cdots, \p_L \} \in \mathbb{R}^3$, where $L$ is the number of points in Cartesian space provided by the motion capture system. We then filter out the bag points $\P^F$ by removing outliers based on their relative position, and identify the points $\P^R \in \P^F$ belonging to the rim\footnote{For more details see our implementation \url{https://sites.google.com/view/bilbo-bag}.}.

The optimal state of an opened bag is characterized by a large, round rim and an expanded volume, which is required for downstream tasks such as item insertion. To quantify this state, we introduce three bag metrics that measure the bag volume, rim elongation, and rim area. These metrics utilize a convex hull function $CH(\cdot)$ of the bag points.
The bag volume is defined as the 3D convex hull of the filtered points
\begin{equation}
    V = \text{Volume}(\text{CH}_\text{3D}(\P^F)).
\end{equation}
To measure the elongation, we first extract the \ac{pca} axes from the 2D convex hull of the rim points
\begin{equation}
    \lambda_1, \lambda_2, \mathbf{v}_1, \mathbf{v}_2  = \text{PCA}( \text{Vertices}(\text{CH}_\text{2D}(\P^R_{x,y}))).
\end{equation}
Note that we use subscripts to denote that values for specific coordinates are extracted throughout this section.
The elongation metric is then defined as the ratio of the \ac{pca} axes so that the length of the axis primarily directed along the y-axis of the robot frame is divided by the length of the axis primarily directed along the x-axis 
(see Fig. \ref{fig:full_pipeline} for the axes)
\begin{equation}
    E = 
    \begin{cases} \sqrt{\frac{\lambda_2}{\lambda_1}} & \text{if} \; | \mathbf{v}_{1,x}| > |\mathbf{v}_{2,x}| , \\
    \sqrt{\frac{\lambda_1}{\lambda_2}} & \text{otherwise}. \label{eq:Elongation}
    \end{cases}
\end{equation}
This definition of the elongation metric is beneficial for the refinement of the elongation using the robot system, in contrast to defining the elongation as the ratio between the major and minor \ac{pca} axes~\cite{chen2023autobag} which does not take into account the direction of the elongation.

\begin{figure}[t]
\vspace{0.2cm}
\centering
\setlength{\belowcaptionskip}{-10pt}
\def\svgwidth{0.9\linewidth}
{\fontsize{7}{7}
%% Creator: Inkscape 1.3.2 (091e20e, 2023-11-25), www.inkscape.org
%% Accompanies image file 'area_metric.pdf' (pdf, eps, ps)
\begingroup%
  \makeatletter%
  \providecommand\color[2][]{%
    \errmessage{(Inkscape) Color is used for the text in Inkscape, but the package 'color.sty' is not loaded}%
    \renewcommand\color[2][]{}%
  }%
  \providecommand\transparent[1]{%
    \errmessage{(Inkscape) Transparency is used (non-zero) for the text in Inkscape, but the package 'transparent.sty' is not loaded}%
    \renewcommand\transparent[1]{}%
  }%
  \providecommand\rotatebox[2]{#2}%
  \newcommand*\fsize{\dimexpr\f@size pt\relax}%
  \newcommand*\lineheight[1]{\fontsize{\fsize}{#1\fsize}\selectfont}%
  \ifx\svgwidth\undefined%
    \setlength{\unitlength}{663.7673013bp}%
    \ifx\svgscale\undefined%
      \relax%
    \else%
      \setlength{\unitlength}{\unitlength * \real{\svgscale}}%
    \fi%
  \else%
    \setlength{\unitlength}{\svgwidth}%
  \fi%
  \global\let\svgwidth\undefined%
  \global\let\svgscale\undefined%
  \makeatother%
  \begin{picture}(1,0.25503525)%
    \lineheight{1}%
    \setlength\tabcolsep{0pt}%
    \put(0,0){\includegraphics[width=\unitlength,page=1]{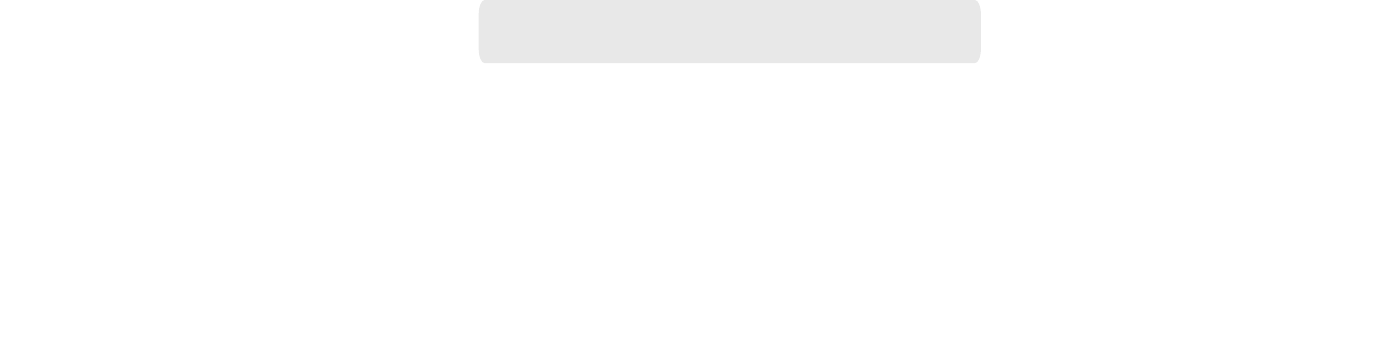}}%
    \put(0.5884211,0.2235278){\color[rgb]{0,0,0}\makebox(0,0)[lt]{\lineheight{1.25}\smash{\begin{tabular}[t]{l}$\alpha$-Shape\end{tabular}}}}%
    \put(0,0){\includegraphics[width=\unitlength,page=2]{area_metric.pdf}}%
    \put(0.38319986,0.22370656){\color[rgb]{0,0,0}\makebox(0,0)[lt]{\lineheight{1.25}\smash{\begin{tabular}[t]{l}Convex Hull\end{tabular}}}}%
    \put(0,0){\includegraphics[width=\unitlength,page=3]{area_metric.pdf}}%
    \put(0.00284776,0.01174292){\color[rgb]{0,0,0}\makebox(0,0)[lt]{\lineheight{1.25}\smash{\begin{tabular}[t]{l}a)\end{tabular}}}}%
    \put(0.50457283,0.01525514){\color[rgb]{0,0,0}\makebox(0,0)[lt]{\lineheight{1.25}\smash{\begin{tabular}[t]{l}b)\end{tabular}}}}%
    \put(0,0){\includegraphics[width=\unitlength,page=4]{area_metric.pdf}}%
  \end{picture}%
\endgroup%
}
\caption{Estimation of the opening area using the $\alpha$-shape (orange), and convex hull (black). In a) the convex hull overestimates the rim area. In b) the $\alpha$-shape underestimates the area due to the occlusion of markers.} 
\label{fig:alphaTradeoffs}
\end{figure}

Finally, the area metric is defined by fitting an alpha-shape \cite{OriginalalphaShapes} to the rim points
\begin{equation}
    A = \text{Area}(\text{AlphaShape}(\P^R_{x,y},\alpha)).
\end{equation}
To accurately define multiple bag openings we need different $\alpha$ values.
Thus, we define $\alpha$ as a linear function of the 2D convex hull of the rim
\begin{equation}
    \alpha = k_{\alpha}*\text{Area}(\text{CH}_\text{2D}(\P^R_{x,y})) + b_{\alpha},
\end{equation}
where $k_\alpha$ and $b_\alpha$ were found empirically for bags of different sizes. Our definition of the rim area provides better estimates than directly using the convex hull~\cite{chen2023autobag}, which can lead to an overestimation of the area when the rim is slim and bent (see Fig. \ref{fig:alphaTradeoffs} a). However, alpha-shapes can underestimate the opening area for some bag configurations, e.g., when points of the bag are occluded (see Fig. \ref{fig:alphaTradeoffs} b).
Nevertheless, overestimating the area is a more severe problem as it is prone to early task termination in states which are actually insufficient, whereas underestimating the area only results superfluous motions for improving the area metric.

\subsection{Elongation Refinement Motion}
\label{sec: BILBO Refinement Motion}

% Alternative (not sure about it)
The dynamic motion from the human demonstrator targets only a high volume and rim area, without refining the roundness of the opening.
Consequently, as suggested in Section \ref{sec: FRAMEWORK Refinement Motion} we design a cost function on the elongation metric defined in equation \eqref{eq:Elongation}.
The optimal elongation $E=1$ indicates a perfectly round opening and stretching in any direction is undesirable. Therefore, we define the cost function as the distance to optimal elongation: $\Delta\text{Elongation} =  C(\s) = |1-E(\s)|$.

As actions, we use linear, reversible motions that adjust the distance between the grippers along the x-axis by a pre-defined step. 
We implement the action choice in equation \eqref{eq:argmin_cost}  with a state-machine based on the relationship between the gripper distance and the elongation of the bag.
In case that $E<1$, the bag is elongated primarily along the x-axis of the robot frame, and the refinement will decrease the distance between the grippers.
On the other hand, if $E>1$, the bag is elongated along the y-axis and the motion will increase the grippers' distance.
Additionally, to prevent collisions and tearing apart the bag, we defined a maximum and minimum allowed distance along the x-axis.

\section{{Experimental Set-up}}
\label{sec:exp_and_res}
Our experiments are designed to answer the following research questions:

\begin{itemize}
    \item What is the performance of different constrained \acp{dmp} for dynamic manipulation of deformable objects?   
    \item Does including a quasi-static refinement stage improve the manipulation performance?
    \item How well does BILBO generalize to bags that are significantly larger than the one used to record the human demonstration?
    \item How does the difficulty of opening a bag relate to the size and stiffness of the material?
\end{itemize}

Our dual-arm setup consists of two Franka Emika's robots\footnote{We used two different versions of the Franka Emika's robot, Panda and Franka Research 3, due to hardware resources.} and an OptiTrack motion capture system for perception. Each experiment was repeated 10 times for each combination of bags, manipulation approach, and initial bag configuration. In total, 240 evaluation runs were gathered in the experiments.

For the constrained \ac{dmp} implementations, we set the constraints to 98\% of the strictest position, velocity, and acceleration limits for each joint
specified in \cite{frankaInterfaceSpec} to leave a margin for numerical inaccuracies.
Both $\tau$ of tau-\ac{dmp} and $\gamma_a$ of TC-\ac{dmp}, a parameter which balances the trade-off between avoiding limit violations and increasing slowdown, were set by gradually increasing their values until no limits were exceeded.
In Opt-\ac{dmp}, a parameter $\lambda$ is used to toggle between optimizing with respect to position and velocity.
In our case, we optimize with respect to the position to reduce the distortion from the demonstrated trajectory, which is verified to be safe, and therefore lower the risk of collision between the robots.
All kinematic constraints are enforced regardless of the optimization objective. 
To ensure a fair comparison between the \ac{dmp} methods, we tuned the parameters so that they produced comparable trajectories when no constraints were applied.

\subsection{Bags and Initial Configuration}
\label{sec:bags}
\begin{figure}[t]
\centering
\def\svgwidth{\linewidth}
{\fontsize{6}{6}%\selectfont\sf
\input{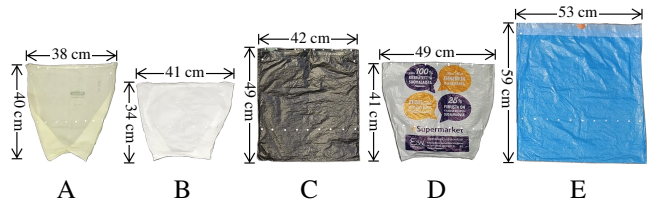}}
\setlength{\abovecaptionskip}{-8pt}
\setlength{\belowcaptionskip}{-16pt}
\caption{Plastic bags used in the experiments.} 
\label{fig:bags}
\end{figure}

In our experiments, we used five plastic bags with different material properties and sizes (see Fig.~\ref{fig:bags}). To provide context on the size of the evaluated bags, prior work \cite{chen2023autobag,chen2023SLIPbagging} has demonstrated quasi-static manipulation of bags in the ranges 28-30 cm by 49-54 cm and 32-55 cm by 32-55 cm, respectively, while \cite{gu2023shakingbot} dynamically manipulated bags in the range of 25-35 cm by 40-53 cm.

Regarding the material, bag A is made of biodegradable plastic, making it especially soft, while the other bags are made of polyethylene. Moreover, bags C and E contain drawstrings, which increases the rigidity of the rim.
The stiffness of the bags, as perceived by a human evaluator, listed from soft to stiff, is as follows: A, C, E, D, B. Note that only bag A was used to record the human demonstration.

\begin{figure}[t]
    \vspace{0.2cm}
     \centering
     \begin{subfigure}[b]{0.23\columnwidth}
         \centering
         \includegraphics[width=\columnwidth]{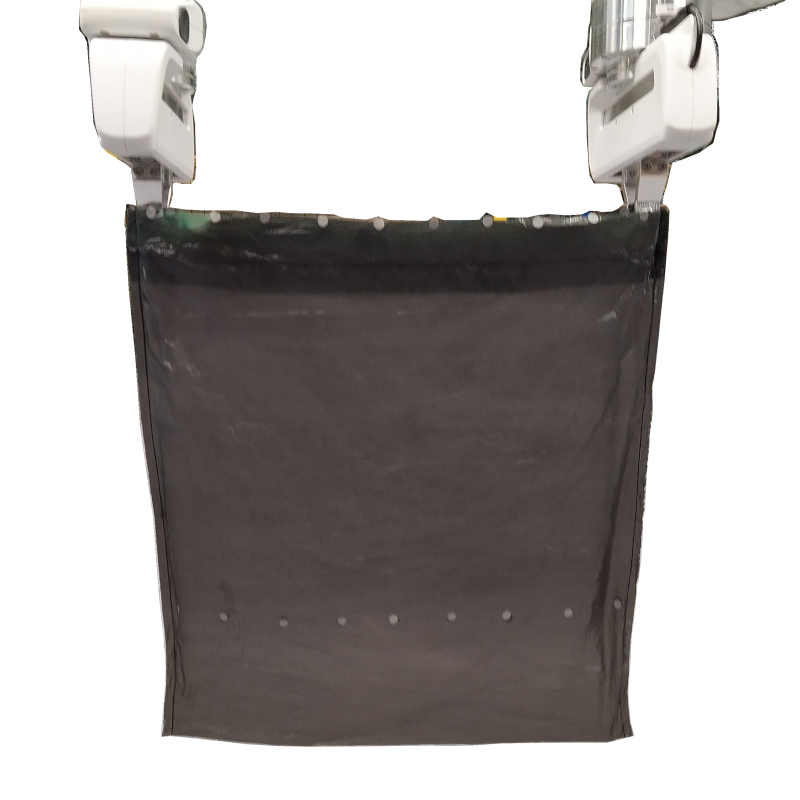}
         \setlength{\abovecaptionskip}{-13pt}
         \setlength{\belowcaptionskip}{0pt}
         \caption{Bag C Easy}
         \label{fig:Bag_C_Easy}
     \end{subfigure}
     \hspace{0.05\columnwidth}
     \begin{subfigure}[b]{0.23\columnwidth}
         \centering
         \includegraphics[width=\columnwidth]{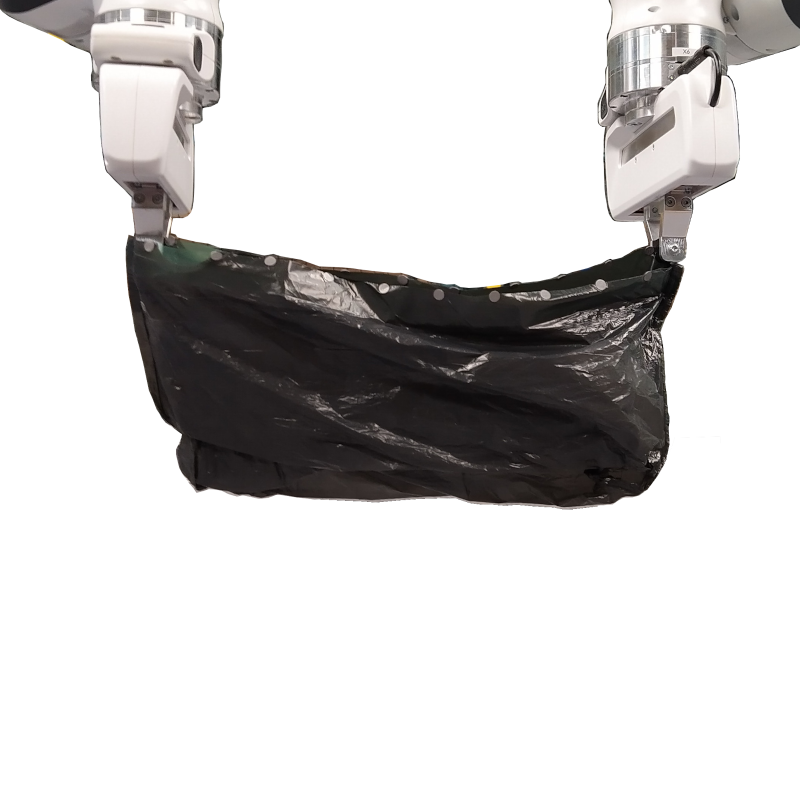}
         \setlength{\abovecaptionskip}{-13pt}
         \setlength{\belowcaptionskip}{0pt}
         \caption{Bag C Hard}
         \label{fig:Bag_C_Hard}
     \end{subfigure}
     \hspace{0.05\columnwidth}
     \begin{subfigure}[b]{0.23\columnwidth}
         \centering
         \includegraphics[width=\columnwidth]{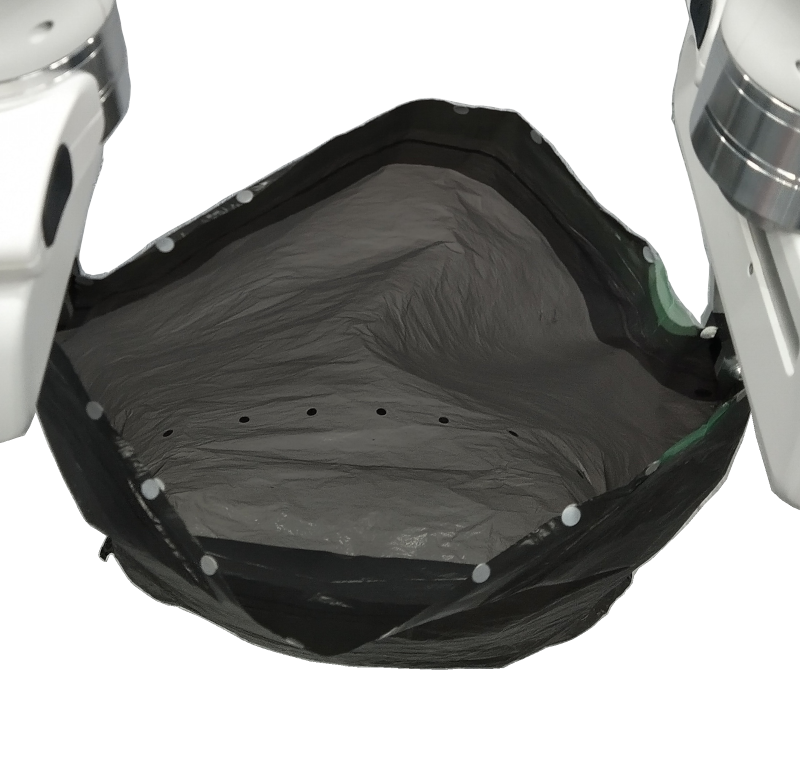}
         \setlength{\abovecaptionskip}{-13pt}
         \setlength{\belowcaptionskip}{0pt}
         \caption{Bag C Open}
         \label{fig:Bag_C_Open}
     \end{subfigure}
        \setlength{\abovecaptionskip}{3pt}
        \setlength{\belowcaptionskip}{-0pt}
        \caption{Examples of two initial bag configurations used in the experiments and an example the final state after BILBO.}
        \label{fig:BagC_states}
\end{figure}

We define two initial states of the bags to evaluate the task performance, an easy state and a hard state (see Fig.~\ref{fig:BagC_states}). In both cases, we assume that the bags are gripped symmetrically by the robots.
In the easy state, the bag is hanging extended with the rim closed.
The hard state has the bottom of the bag crumpled, folded inward, and the rim pressed shut.

\begin{figure*}[t]
\centering
\def\svgwidth{\linewidth}
{\fontsize{6}{6}%\selectfont\sf
\input{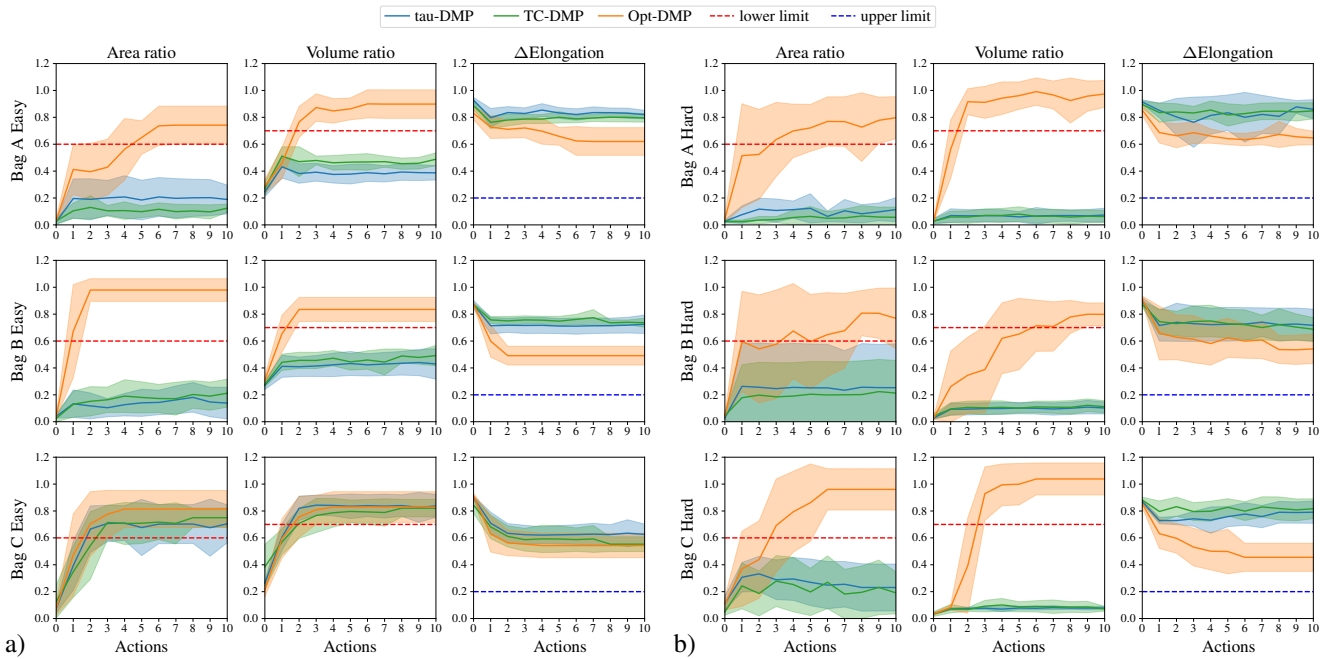}}
\setlength{\abovecaptionskip}{-5pt}
\setlength{\belowcaptionskip}{-15pt}
\caption{Quantitative results of the learned dynamic primitives using Opt-DMP, tau-DMP, and TC-DMP for three different bags in a) easy and b) hard initial configuration. The results show the area, volume, elongation and their target limits.}
\label{fig:fling_only}
\end{figure*}

\subsection{Bag Opening Evaluation}
\label{sec:metrics_eval}

In our experiments, we evaluated the bag-opening task using the bag metrics proposed in Section~\ref{sec: BagMetrics}.
The area and volume metrics are represented as the ratio between the achieved value and the observed value in a successful human demonstration with the same bag.
The area and volume are considered sufficiently high if they reach the lower limits of 60\% and 70\%, respectively, where we consider a bag to be open if these targets are reached.
We use a more strict volume target because high volumes were expected to be easier to achieve and maintain compared to high areas of the more unstable rim.

In addition, we study the elongation during the refinement stage via the cost function defined in Section \ref{sec: BILBO Refinement Motion}. 
In the experiments, we use an upper limit for the cost function, which is deemed sufficiently low when $\Delta\text{Elongation} \leq 0.2$.
This relatively strict upper limit is selected so that the manipulation would not terminate early at an easier target value, and the system would instead use any remaining actions to strive for a better rim elongation.

The refinement action is reversed if it causes the area or volume metrics to fall below the target levels, and in such cases, the refinement will terminate after the reversal.
The reason for this rule is that the area and volume, are treated as primary objectives, which should not be compromised for better elongation, i.e., the secondary objective.
The manipulation scheme terminates if the target elongation is met without reducing the area or volume metrics below their targets, or if the total number of allocated actions is reached.

\section{{Experimental Results}}
\label{sec:results}

\subsection{%Experiment 1: fling-only primitive
Analysis of Constrained DMP methods
} 
\label{sec: fling_only_experiment}

In this experiment, we studied the performance of the dynamic manipulation primitive learned with each of the constrained DMP versions.
We evaluated bags A, B, and C in both easy and hard initial configurations.
The dynamic motion was repeated until the targets set for the area and volume metrics were achieved or a maximum of 10 actions was reached.
Note that runs that needed less than 10 actions to reach the targets have the unused steps padded with the final values in the figures.

As shown in Fig.~\ref{fig:fling_only}, Opt-DMP consistently exceeds the area and volume targets for each bag, while other methods only succeeded in reaching the targets for bag C from the easy initial state.
Although the hard initial state typically requires more actions compared to the easy state, Opt-DMP was able to frequently achieve the targets with only 2-6 actions, depending on the bag and the difficulty of the initial state. This demonstrates the effectiveness of the dynamic motion. As the performance of TC-DMP is comparable to tau-DMP, the only benefit of TC-DMP in the setting we consider is faster runtime, as $\tau$ is adaptive rather than fixed.

A key reason for the superior performance of Opt-DMP compared to the $\tau$-scaling methods is that Opt-DMP enables high peak velocities and accelerations in each joint independently, at the cost of distorting the path, as the constraints are encoded in the DMP weights.
In contrast, the tau-DMP and TC-DMP tend to slow down all joints via the shared canonical system, in case any joint exceeds the limits, as they enforce the constraints by scaling $\tau$.
This result highlights that the ability of Opt-DMP to independently gain high peak velocities and accelerations in each joint is more important for dynamically manipulating the bag than an accurate reproduction of the demonstrated path.

In addition, the results show the relationship between the structure and stiffness of the bag and its opening performance. Bag A, which was used to provide the human demonstration, ended up being relatively difficult to open as the rim would easily collapse due to the softness of the material.
Due to the stiffness of bag B, it needed relatively many repetitions of the dynamic motion to uncrumple the bag initialized in the hard state and reach the target volume.
In contrast, the medium stiffness and drawstring in the rim of bag C made it the easiest bag to open in this experiment, despite its significant depth compared to bags A and B.

Note that the dynamic motion can only slightly enhance the elongation.
However, the final elongation is far from the target value.
This motivates the proposed refinement motion, since further refinement of the elongation is necessary.

\subsection{%Experiment 2: refinement-only motion
Evaluating the Elongation Refinement Motion} 
\label{sec: refine_only_experiment}

In this experiment, we studied the performance of the refinement motion for improving the elongation metric. 
In addition, we evaluated the bag-opening performance of the refinement motion alone to rule out the possibility that it alone is sufficient without dynamic manipulation.
In this experiment, bag A was initialized in the hard state, where the refinement motion was applied until a maximum of 20 actions were performed or the area and volume targets were reached.
While BILBO simply prevents motion outside a range we defined to prevent collisions or the bags tearing, in this experiment we had the robots take a step in the opposite
direction if they were commanded past these limits. This
allowed the robots to keep adjusting the bag, and potentially
uncrumple it, instead of terminating the run.

\begin{figure}[t]
\centering
\def\svgwidth{\linewidth}
{\fontsize{6}{6}%\selectfont\sf
\input{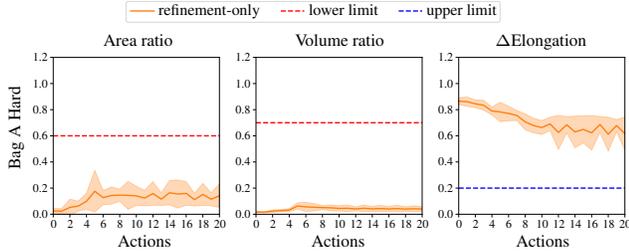}}
\setlength{\abovecaptionskip}{-9pt}
\setlength{\belowcaptionskip}{-18pt}
\caption{
Area, volume and elongation results using only the refinement motion for bag A initialized in the hard state. %The limits defined for each metric are also plotted.
}
\label{fig:refine_only}
\end{figure}

The results in Fig.~\ref{fig:refine_only} show that the refinement motion can slightly improve the elongation, suggesting that the motion can be used to improve the roundness of the bag opening.
However, the refinement motion neither manages to uncrumple the bag nor approaches the area or volume limits, which highlights the necessity of using the dynamic primitive.
Therefore, the optimal system should combine the best-performing dynamic primitive obtained from the Opt-DMP with the non-dynamic refinement scheme.

\subsection{%Experiment 3: BILBO
BILBO's Bag-Opening and Generalization Results} 
\label{sec: full_AC_experiment}

Finally, we evaluated the performance of our proposed system BILBO, which combines the dynamic primitive encoded by Opt-DMP with the linear refinement.
First, we evaluated the performance of the system using bags A, B, and C, which have similar widths and varying materials.
A maximum of 20 total actions were allocated for both the dynamic and refinement motions. 
The dynamic primitive was first applied until the target area and volume were reached. Then, if the maximum number of actions had not been reached, the refinement primitive would be subsequently applied based on the measured elongation. An example of the final state of a successful run of BILBO is shown in Fig. \ref{fig:BagC_states} c.

\begin{figure}[t]
\centering
\def\svgwidth{\linewidth}
{\fontsize{6}{6}%\selectfont\sf
\input{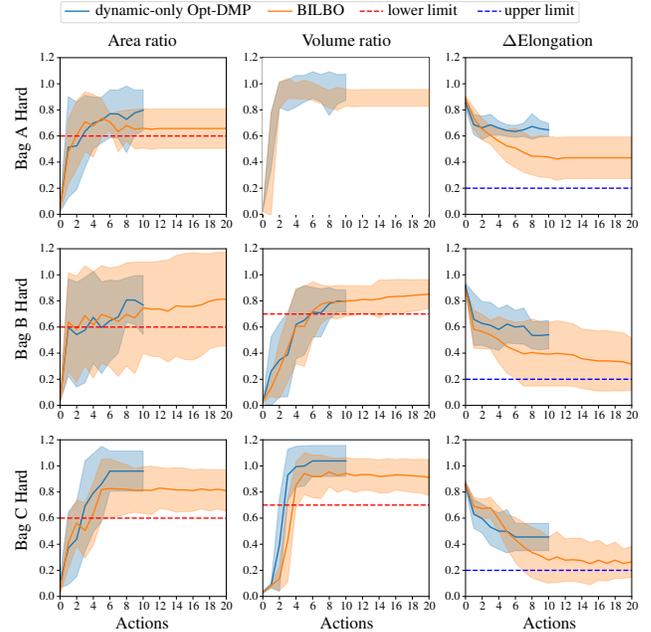}}
\setlength{\abovecaptionskip}{-11pt}
\setlength{\belowcaptionskip}{-14pt}
\caption{
Area, volume, and elongation results of BILBO for bags A-C initialized in the hard state. The results from the dynamic primitive experiment are also shown as a reference, denoted as dynamic-only Opt-DMP.} 
\label{fig:full_AC}
\end{figure}

The performance of both BILBO and Opt-DMP from the dynamic primitive experiment is shown in Fig.~\ref{fig:full_AC}.
The results show that BILBO's refinement motion is able to keep the target state achieved by the dynamic motion, as the area and velocity targets are reached at a level similar to that of the dynamic-only experiment, even for bag A.
Furthermore, the refinement successfully decreases the distance to optimal elongation.
The target elongation is more challenging to reach for bag A, as the soft material causes the rim to collapse easily.
In contrast, it is not uncommon for bags B and C to reach the target.
One thing to note is that the mean values of  $\Delta\text{Elongation}$ do not fall below the target level for any bag.
The reason for this is that the target was intentionally chosen as a strict value to encourage finding the best reachable elongation as explained in Section \ref{sec:metrics_eval}. Therefore, $\Delta\text{Elongation}$ converges to the best reachable level for each bag in our experiments.

\begin{figure}[t]
\centering
\def\svgwidth{\linewidth}
{\fontsize{6}{6}%\selectfont\sf
\input{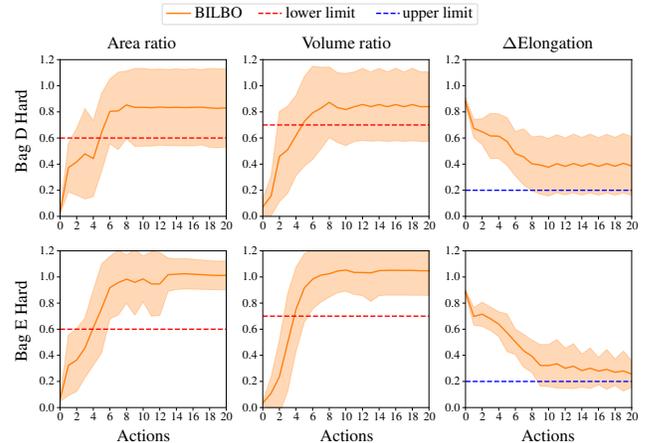}}
\setlength{\abovecaptionskip}{-6pt}
\setlength{\belowcaptionskip}{-20pt}
\caption{Area, volume, and elongation results of BILBO for bags D \& E initialized in the hard state.} 
\label{fig:full_DE}
% \vspace{-7mm}
\end{figure}

Additionally, we evaluated BILBO's capability to generalize to significantly larger bags than the one used for demonstration by conducting the same experiment on Bag D and Bag E. 
The results in Fig.~\ref{fig:full_DE} show that BILBO yields similar performance on bags D and E as it does on bags A-C. 
Even with significantly larger bags, BILBO consistently surpasses the target values for both area and volume metrics, maintaining stability during refinement, while the $\Delta\text{Elongation}$ converges to a bag-specific lowest level.

All in all, the results show that BILBO is capable of successfully opening different bags that vary in size and material with just one human demonstration for a single bag, highlighting both the generalizability and efficiency of the proposed method.
One possible extension is to extend BILBO to the full bagging task, including gripping and item insertion, thus developing the \textit{BILBO Bagging} system.

\section{Conclusions}
\label{sec:conclusions}
We presented an \ac{il} framework that first learns a dynamic primitive for inducing significant changes in the states of deformable objects with few motions, while adhering to robot constraints with constrained \acp{dmp}. Subsequently, it employs stable quasi-static motions for small-scale adjustment.
We assessed the framework in the task of bimanual bag opening, resulting in a system named BILBO. 

First, we evaluated three constrained \ac{dmp} versions for generating a dynamic primitive that adheres to the robots' limits. Our experiments showed that Opt-DMP achieved superior performance, which demonstrates that in the dynamic manipulation task, it is more important to achieve high velocities and accelerations than to accurately reproduce the demonstrated path. This highlights the necessity of selecting a constraint satisfaction approach that prioritizes attributes most relevant to the task.
Due to the benefits of Opt-DMP, it was selected for learning the dynamic motion in the BILBO system using a human demonstration with a single bag. Our experimental results showed that BILBO was able to effectively open bags of different materials, shapes, and sizes than the one used for the demonstration.
More specifically, BILBO was able to open most bags in less than 5 actions and achieve near-optimal elongation after 20 actions.

As future directions, the learned dynamic motions could be used to bootstrap learning methods that refine the motion by, for example, adapting the \ac{dmp} weights, without violating the kinematic constraints. This could potentially lead to greater efficiency in dynamic manipulation tasks.

%%%%%%%%%%%%%%%%%%%%%%%%%%%%%%%%%%%%%%%%%%%%%%%%%%%%%%%%%%%%%%%%%%%%%%%%%%%%%%%%

% \section*{Acknowledgements}

\bibliographystyle{IEEEtran}
\bibliography{refs}

\end{document}